\documentclass[10pt,a4paper]{article}

\usepackage[utf8]{inputenc}
\usepackage[T1]{fontenc}
\usepackage{lmodern}
\usepackage[margin=1.75cm,top=1.85cm,bottom=1.95cm]{geometry}
\usepackage{amsmath,amssymb,amsfonts,amsthm}
\usepackage{booktabs}
\usepackage{array}
\usepackage{graphicx}
\usepackage{xcolor}
\usepackage{hyperref}
\usepackage{enumitem}
\usepackage{fancyhdr}
\usepackage{titlesec}
\usepackage{caption}
\usepackage{listings}

\definecolor{accentblue}{HTML}{0F3D6E}
\definecolor{linkblue}{HTML}{1A5276}
\definecolor{boxgray}{HTML}{F4F6F8}
\definecolor{codebg}{HTML}{F8FAFC}
\definecolor{codeframe}{HTML}{CBD5E1}

\hypersetup{
    colorlinks=true,
    linkcolor=linkblue,
    citecolor=linkblue,
    urlcolor=linkblue,
    pdftitle={Zero-Storage Procedural Neural Synthesis via Boundary Dynamics: Formal Verification in Lean 4 and Bare-Metal Gauntlet Validation (v2)},
    pdfauthor={Volkan Dagli, Zerrin Dagli, Daghan Dagli}
}

\fancypagestyle{first}{
    \fancyhf{}
    \fancyfoot[C]{\small\thepage}
    
}

\titleformat{\section}{\large\bfseries\color{accentblue}}{\thesection.}{0.5em}{}
\titleformat{\subsection}{\normalsize\bfseries\color{accentblue}}{\thesubsection}{0.5em}{}
\titlespacing*{\section}{0pt}{2.2ex plus 0.5ex minus 0.2ex}{1.0ex}
\titlespacing*{\subsection}{0pt}{1.6ex plus 0.4ex minus 0.2ex}{0.6ex}

\lstdefinelanguage{Lean4}{
  keywords={import, namespace, end, def, theorem, lemma, by, decide, omega, simp, dsimp, rcases, with, induction, generalizing, zero, succ, split, exact, apply, constructor, nlinarith, let, if, then, else, instance, fun, where, set_option},
  sensitive=true,
  comment=[l]{--},
  morecomment=[s]{/-}{-/},
  basicstyle=\ttfamily\scriptsize,
  keywordstyle=\bfseries\color{accentblue},
  commentstyle=\itshape\color{gray!70!black},
  backgroundcolor=\color{codebg},
  frame=single,
  rulecolor=\color{codeframe},
  breaklines=true,
  showstringspaces=false,
  tabsize=2,
  xleftmargin=4pt,
  xrightmargin=4pt,
  aboveskip=6pt,
  belowskip=6pt,
  literate=
    {ℕ}{{$\mathbb{N}$}}1
    {ℤ}{{$\mathbb{Z}$}}1
    {∀}{{$\forall$}}1
    {→}{{$\to$}}1
    {∨}{{$\lor$}}1
    {¬}{{$\lnot$}}1
    {∧}{{$\land$}}1
    {≤}{{$\le$}}1
    {≥}{{$\ge$}}1
    {≠}{{$\neq$}}1
    {⟨}{{$\langle$}}1
    {⟩}{{$\rangle$}}1
    {↑}{{$\uparrow$}}1
    {·}{{$\cdot$}}1
    {×}{{$\times$}}1
}

\begin{document}
\thispagestyle{first}

\begin{center}
{\small\bfseries\color{accentblue} RESEARCH MONOGRAPH \,$\cdot$\, PREPRINT VERSION 2 \,$\cdot$\, 26 SEPTEMBER 2026}\\[1.2ex]
{\LARGE\bfseries Zero-Storage Procedural Neural Synthesis via\\[0.3ex] Boundary Dynamics: Formal Verification in Lean 4\\[0.3ex] and Bare-Metal Gauntlet Validation\par}
\vspace{1.8ex}
{\large \textbf{Volkan Dağlı}\textsuperscript{1,2} \quad \textbf{Zerrin Dağlı}\textsuperscript{3} \quad \textbf{Dağhan Dağlı}\textsuperscript{4}}\\[1.0ex]
{\small
\textsuperscript{1}Anadolu University, Türkiye \quad
\textsuperscript{2}ITouch Systems, Türkiye \quad
\textsuperscript{3}Mersin University, Türkiye \quad
\textsuperscript{4}Toros Science College, Türkiye\\[0.8ex]
ORCID: V.~Dağlı \href{https://orcid.org/0009-0000-1587-8703}{0009-0000-1587-8703} \,$\cdot$\, Z.~Dağlı \href{https://orcid.org/0000-0001-9490-6425}{0000-0001-9490-6425} \,$\cdot$\, D.~Dağlı \href{https://orcid.org/0009-0003-2492-8313}{0009-0003-2492-8313}}
\end{center}

\vspace{0.5ex}
\noindent\colorbox{boxgray}{\parbox{\dimexpr\linewidth-2\fboxsep}{\footnotesize
\textbf{Author Note (Version 2 Revision):} This revised monograph supersedes Version~1 (\texttt{doi:10.5281/zenodo.22974544}). Version~2 substantially expands the interactive Lean~4 formal verification suite (\texttt{WerracleProof.lean}, compiled under Lean \texttt{v4.34.1} and Mathlib4 with zero \texttt{sorry} axioms) to include: (i) explicit algebraic proofs of additive closure, multiplicative ideal absorption, and orthogonal state partitioning over \texttt{ZMod 9} ($\mathbb{Z}/9\mathbb{Z}$); (ii) 64/128-bit $\mathbb{Q}_{16.16}$ fixed-point arithmetic overflow-safety bounds; (iii) formal refutation of constant-output degeneracy via machine-verified escape spectrum and 4-quadrant Observer Horizon sensitivity theorems; and (iv) a parametric Ethereum Virtual Machine (EVM) gas cost model bounding worst-case execution to $22,557 \le 24,000$ gas. Version~2 also formalizes the two-tier execution architecture linking the 36-iteration CPU edge engine (\texttt{arXiv:2609.25498}, \texttt{arXiv:2609.30115}) with the 12-iteration on-chain EVM reflex kernel, harmonizes all 40-core benchmark tables, reports explicit false-positive / clean-signal preservation rates ($89.60\%$), and provides a complete open-source replication archive.}}

\vspace{0.8ex}

\begin{abstract}
\noindent Contemporary neural inference architectures rely on dense floating-point weight matrices stored in high-bandwidth memory (VRAM), incurring severe memory-wall bottlenecks and preventing native execution inside deterministic, resource-constrained virtual machines such as the Ethereum Virtual Machine (EVM). While formal verification of finite feedforward ReLU networks is well established as NP-complete via SMT and branch-and-bound solvers (e.g., Reluplex, Marabou, $\alpha,\beta$-CROWN), verifying termination and arithmetic invariants for unbounded recursive dynamical systems over continuous domains is generally undecidable in the Blum--Shub--Smale model. Here, we present the formal verification and bare-metal empirical validation of \textbf{WERR (Waves \& Errors)} and \textbf{Phase~III Orbital Error Dynamics (OED)}, a non-tensor decision paradigm that procedurally synthesizes non-linear decision boundaries on demand from a 24-byte coordinate seed $\Theta = (c_x, c_y, \text{zoom})$ along the boundary of the Mandelbrot set ($\partial\mathcal{M}$). By projecting the quadratic recurrence $z_{n+1} = z_n^2 + c$ onto the modular residue ring $\mathbb{Z}/9\mathbb{Z}$ and the fixed-point domain $\mathbb{Q}_{16.16}$, we establish ten machine-verified theorems in \textbf{Lean~4 (v4.34.1)} with Mathlib4 and zero unproven conjectures (\texttt{sorry}): proving $\mathcal{I}_3 = \{0,3,6\} \subset \mathbb{Z}/9\mathbb{Z}$ ideal closure, universal fuel-bounded halting ($\le 9$ and $\le 12$ steps), absence of $\mathbb{Q}_{16.16}$ intermediate square overflow below $2^{63}-1$, non-constant boundary escape sensitivity across 4 quadrants, and a parametric EVM gas bound ($\le 22,557 \le 24,000$ gas). Evaluated on a bare-metal 40-core Dual Intel Xeon E5-2630 v4 server (256~GB ECC RAM), the vectorized 36-iteration CPU kernel processes 100,000 parallel decisions in $6.49\text{ s}$ ($15,397.4\text{ decisions/s}$, median latency $2.349\text{ ms}$, $0\text{ Bytes}$ persistent tensor VRAM, $15.15\times$ speedup over an unvectorized 40-core baseline). Furthermore, a heavy-tailed Cauchy L\'{e}vy-flight saddle escape operator achieves an $86.90\%$ escape rate ($R \ge 0.08$, $20.22\ \mu\text{s/trial}$), a sigmoidal outlier attenuation gate ($M_{\text{CD4}}$) achieves $100.00\%$ adversarial spike suppression while preserving $89.60\%$ of clean baseline signals at $8.59 \times 10^6\text{ packets/s}$, and a 3-arm ablation over 180 semi-primes ($40$--$56$ bits) delineates the exact operational boundary between discrete cycle-finding rings ($\mathbb{Z}/N\mathbb{Z}$) and continuous topological manifolds.

\vspace{0.6ex}
\end{abstract}

\section{Introduction and Prior Art}

Modern deep learning architectures parameterize decision boundaries through millions or billions of static floating-point weights stored in volatile memory (DRAM/VRAM). During inference, shuttling these parameters between memory and arithmetic logic units creates the classic Von Neumann memory wall \cite{wulf1995}, resulting in high latency and substantial energy dissipation \cite{vaswani2017, achiam2023}.

In parallel, deploying learned decision policies inside safety-critical edge controllers or decentralized state machines---specifically the Ethereum Virtual Machine (EVM)---faces three structural barriers:
\begin{enumerate}[leftmargin=1.6em,itemsep=1pt,topsep=2pt]
    \item \textbf{Storage Cost Prohibitions:} Storing even a modest $10^5$-parameter 32-bit weight matrix in EVM persistent storage (\texttt{SSTORE} at $20,000$ gas per 32-byte word) requires tens of millions of gas, exceeding block limits.
    \item \textbf{Absence of Native Floating-Point Opcodes:} While basic IEEE~754 arithmetic operations are deterministic under fixed rounding modes, cross-platform differences in transcendental approximations and non-associative parallel reductions preclude bit-exact consensus, and the EVM natively supports only 256-bit integer arithmetic.
    \item \textbf{Proving Latency in Zero-Knowledge ML (ZK-ML):} Off-chain cryptographic frameworks (e.g., EZKL, Halo2-based SNARKs) verify neural inference off-chain but incur $10$--$300\text{ s}$ of proof generation latency and $250,000$--$500,000$ verification gas, preventing atomic, intra-block circuit breaking against flash-loan or sandwich exploits.
\end{enumerate}

\subsection{Formal Verification of Neural Systems vs.\ Recursive Dynamical Kernels}
A rich literature addresses the formal verification of finite feedforward neural networks. Seminal work by Katz et al.\ \cite{katz2017} (Reluplex) established that verifying piecewise-linear ReLU networks is decidable and NP-complete. Subsequent frameworks including Marabou \cite{katz2019} and GPU-accelerated bound-propagation verifiers such as $\alpha,\beta$-CROWN \cite{wang2021} verify local adversarial robustness across large finite-depth networks. Because a standard feedforward neural network consists of a fixed acyclic graph of arithmetic operations, its execution termination (halting) is trivial by construction.

By contrast, \textbf{procedural dynamical synthesis} replaces static weight layers with an iterative non-linear recurrence $z_{n+1} = z_n^2 + c$ evaluated near the fractal boundary of the Mandelbrot set $\partial\mathcal{M}$ \cite{dagli2026mandelbrot, dagli2026oed, dagli2026werr}. Over the continuous complex plane $\mathbb{C}$, deciding whether an arbitrary orbit escapes or halts is formally undecidable in the Blum--Shub--Smale model of real computation \cite{blum1989}. Consequently, deploying a recursive fractal decision kernel inside an autonomous smart contract (\texttt{Werracle.sol} \cite{dagli2026werracle}) requires rigorous, machine-checked proofs of:
\begin{enumerate}[leftmargin=1.6em,itemsep=1pt,topsep=2pt]
    \item \textbf{Fuel-Bounded Termination:} Every coordinate evaluation halts within a strict step bound $N_{\max}$.
    \item \textbf{Fixed-Point Overflow Safety:} All intermediate $\mathbb{Q}_{16.16}$ products remain strictly within signed 64/128-bit integer bounds prior to escape truncation.
    \item \textbf{Non-Constant Boundary Sensitivity:} The quantized finite-step kernel produces non-degenerate, input-sensitive quadrant weights across the active boundary locus.
    \item \textbf{Parametric Gas Boundedness:} Worst-case EVM gas consumption remains strictly below the intra-block hook ceiling ($24,000$ gas).
\end{enumerate}

In this paper, we provide the complete interactive formalization of these four properties in the \textbf{Lean~4 theorem prover (v4.34.1)} \cite{moura2021} using Mathlib4, alongside 40-core bare-metal empirical telemetry and a 3-arm number-theoretic boundary ablation.

\section{Two-Tier Procedural Architecture and Mathematical Formulation}

\subsection{Tier 1 (CPU Edge Runtime) vs.\ Tier 2 (On-Chain EVM Reflex Kernel)}
A critical architectural distinction governs how procedural Mandelbrot synthesis scales across hardware targets. While the boundary $\partial\mathcal{M}$ has Hausdorff dimension $D_H = 2$ \cite{shishikura1998}, resolving fine fractal filaments at high magnification requires higher iteration budgets than an on-chain EVM hook can afford. Accordingly, the WERR ecosystem operates as a two-tier hierarchy:
\begin{itemize}[leftmargin=1.6em,itemsep=2pt,topsep=2pt]
    \item \textbf{Tier 1 --- High-Resolution CPU Edge Engine (\texttt{WERR v2.0} \& \texttt{OED} \cite{dagli2026oed, dagli2026werr}):} Evaluates a $32 \times 32$ or $36 \times 36$ spatial grid with iteration ceiling $N_{\max} = 36$ (or $128$ in Phase~I \cite{dagli2026mandelbrot}), boundary-corrected kernel density estimation, and a 3-scale Harmonic Tripod ($0.60\times, 1.00\times, 1.60\times$ zoom). At $N_{\max} = 36$, deep boundary seeds such as the Seahorse Valley locus $c_{\text{deep}} = -0.743643887 + 0.131825904i$ and the Observer Horizon twin shoulders $X_{\text{upper/lower}} = 0.25 \pm 0.18i$ exhibit rich spatial variance across all four quadrants, achieving $99.30\% \pm 0.18\%$ accuracy on the continuous Two-Moons benchmark \cite{dagli2026mandelbrot}, $77.67\% \pm 5.35\%$ under strict 5-seed zero-leakage feedforward evaluation \cite{dagli2026oed}, and $93.1\%$ macro accuracy across 326 multi-domain triage tasks on JevBench v1.4 \cite{dagli2026werr}.
    \item \textbf{Tier 2 --- Quantized On-Chain EVM Reflex Kernel (\texttt{Werracle.sol} \cite{dagli2026werracle}):} Packs the entire model state into a single 32-byte EVM storage slot (\texttt{bytes32}: \texttt{int64 cx}, \texttt{int64 cy}, \texttt{uint64 zoom}, \texttt{uint32 nonce}, \texttt{uint16 threshold}, \texttt{uint8 mode}, \texttt{uint8 activeFlag}) and evaluates a sparse cardinal or 16-point Pareto micro-grid in $\mathbb{Q}_{16.16}$ fixed-point arithmetic with a tight fuel limit $N_{\max} \in \{9, 12\}$.
\end{itemize}

\noindent\textbf{Resolving Shallow-Fuel Degeneracy:} If a 9-step kernel ($N_{\max}=9$) is evaluated at the deep Seahorse Valley center $(-0.7436 + 0.1318i)$ with a narrow window ($2/\text{zoom} \approx 0.04$), no orbit exceeds $|z|^2 > 4$ within 9 steps because local escape times exceed $30$ iterations. Consequently, for shallow fuel budgets $N_{\max} \in \{9, 12\}$, active spatial differentiation occurs along the \emph{outer analytical escape shell} ($|c| \in [0.50, 2.00]$, such as the upper Observer Horizon locus $c = 0.25 + 0.61i$ with step $\Delta = 0.125$), where orbits escape across steps $1, 2, \dots, 9$. In Section~3 (Theorems~4A and 4B), we formally prove in Lean~4 that on this active shell, the 9-step $\mathbb{Q}_{16.16}$ kernel produces four distinct quadrant escape counts $(5, 7, 9, 8)$ and non-zero decision bias.

\subsection{Phase III Orbital Error Dynamics (OED) Operators}
Within the continuous Tier~1 runtime \cite{dagli2026oed}, three analytical mechanisms prevent representation collapse and adversarial divergence:
\begin{enumerate}[leftmargin=1.6em,itemsep=2pt,topsep=2pt]
    \item \textbf{Parabolic Cusp Restorative Drift:} Rather than treating the main cardioid cusp $c_0 = 1/4$ (where the fixed-point multiplier is $\lambda = 1$) as a global overwrite, OED applies a bounded restorative drift term $\nabla E_{\text{drag}}(z) = -k(z - 1/4)$ ($k \in (0,1)$) or samples directly around the sub-boundary Observer Horizon shoulders $c = 0.25 \pm 0.18i$ \cite{dagli2026oed}.
    \item \textbf{Heavy-Tailed Cauchy L\'{e}vy-Flight Escape Operator (Biomimetic ``Zinc-Spark'' Perturbation):} When optimization gradients stagnate in non-convex saddle regions ($\|\nabla f\| < \varepsilon_{\text{stag}} = 0.05$), a classical heavy-tailed Cauchy perturbation $\Omega \sim \text{Cauchy}(0, \gamma)$ with scale $\gamma = 0.12$ is injected into the coordinate state:
    \begin{equation}
        p(\Omega; \gamma) = \frac{1}{\pi \gamma \left[1 + (\Omega/\gamma)^2\right]}, \qquad \Theta_{t+1} = \Theta_t + \Omega \cdot \mathbb{I}_{\{\|\nabla f\| < \varepsilon_{\text{stag}}\}}.
    \end{equation}
    Because the Cauchy distribution exhibits power-law tails $\mathbb{P}(|\Omega| \ge R) = 1 - \frac{2}{\pi}\arctan(R/\gamma)$, a 2D perturbation $(\Omega_x, \Omega_y)$ with $\gamma = 0.12$ exits a saddle basin of radius $R_{\text{saddle}} = 0.08$ in a single step with high analytical probability ($\approx 86.9\%$).
    \item \textbf{Sigmoidal Soft-Thresholding Outlier Gate ($M_{\text{CD4}}$ Regulatory Filter):} To protect streaming sensory inputs $s_t \in \mathbb{R}$ from high-amplitude adversarial spikes without dropping normal traffic, OED applies a smooth sigmoidal attenuation mask (referred to biomimetically in \cite{dagli2026oed} as the $\text{CD4}^+$ tolerance gate):
    \begin{equation}
        M_{\text{CD4}}(s_t) = \frac{1}{1 + \exp\!\big(\gamma_{\text{steep}} \cdot (|s_t| - \tau_{\text{tol}})\big)}, \qquad \tilde{s}_t = s_t \cdot M_{\text{CD4}}(s_t),
    \end{equation}
    with steepness $\gamma_{\text{steep}} = 7.0$ and tolerance threshold $\tau_{\text{tol}} = 0.45$. For clean Gaussian signals $s_t \sim \mathcal{N}(0, 0.15^2)$, $M_{\text{CD4}}(s_t) \approx 0.90$, preserving $89.60\%$ of signal amplitude; for adversarial bursts $s_t \sim \mathcal{U}(2.5, 6.0)$, $M_{\text{CD4}}(s_t) < 10^{-6}$, achieving $100.00\%$ spike suppression.
\end{enumerate}

\subsection{Modular Residue Ring $\mathbb{Z}/9\mathbb{Z}$ and Discrete Phase Aggregation}
In both WERR v2.0 \cite{dagli2026werr} and the Wormhole Error-Kernel framework \cite{dagli2026wormhole}, discrete step counts $k_j \in \mathbb{N}$ are projected into the commutative residue ring $\mathbb{Z}/9\mathbb{Z}$ (`ZMod 9` in Mathlib4). The ring $\mathbb{Z}/9\mathbb{Z}$ contains a unique non-trivial principal ideal generated by $3$:
\begin{equation}
    \mathcal{I}_3 = 3(\mathbb{Z}/9\mathbb{Z}) = \{0, 3, 6\}, \qquad \mathcal{K}_{\text{error}} = (\mathbb{Z}/9\mathbb{Z}) \setminus \mathcal{I}_3 = \{1, 2, 4, 5, 7, 8\}.
\end{equation}
To aggregate $M = 12$ tripod samples without floating-point drift, discrete phase aggregation evaluates the finite root-of-unity sum $\Gamma = \sum_{j=1}^{12} \chi(k_j \bmod 9) e^{2\pi i j / 12}$, where $\chi : \mathbb{Z}/9\mathbb{Z} \to \{-1, 0, +1\}$ vanishes on $\mathcal{I}_3$.

\section{Interactive Formal Verification in Lean 4}

To establish mathematical certainty for on-chain and edge deployment, we formalized the algebraic ring structure, the $\mathbb{Q}_{16.16}$ fixed-point recurrence, the arithmetic overflow bounds, the non-constant boundary sensitivity, and the parametric EVM gas model in \textbf{Lean~4 (v4.34.1)} with \textbf{Mathlib4} (\texttt{rev d13f23b723b8}). The complete proof module \texttt{WerracleProof.lean} compiles cleanly in $2.9\text{ s}$ across 3,093 build targets with \textbf{zero \texttt{sorry} statements} and no custom axioms. Listing~1 presents the verified source code verbatim.

\begin{lstlisting}[language=Lean4,caption={Complete verified Lean 4 proof module (\texttt{WerracleProof.lean}, compiled under Lean \texttt{v4.34.1} + Mathlib4, 0 \texttt{sorry}).}]
import Mathlib.Data.ZMod.Basic
import Mathlib.Data.Int.Basic
import Mathlib.Tactic

set_option linter.style.longLine false

namespace WerracleProof

-- SECTION 1: Algebraic Invariants over the Modular Ring Z/9Z (Mathlib ZMod 9)
def IsResonantSubIdeal (x : ZMod 9) : Prop :=
  x = 0 \/ x = 3 \/ x = 6

instance : DecidablePred IsResonantSubIdeal := fun x =>
  inferInstanceAs (Decidable (x = 0 \/ x = 3 \/ x = 6))

def IsErrorKernel (x : ZMod 9) : Prop :=
  Not (IsResonantSubIdeal x)

instance : DecidablePred IsErrorKernel := fun x =>
  inferInstanceAs (Decidable (Not (IsResonantSubIdeal x)))

theorem zmod9_resonant_additive_closure :
    forall a b : ZMod 9, IsResonantSubIdeal a -> IsResonantSubIdeal b -> IsResonantSubIdeal (a + b) := by
  decide

theorem zmod9_resonant_ideal_absorption :
    forall (r : ZMod 9) (a : ZMod 9), IsResonantSubIdeal a -> IsResonantSubIdeal (r * a) := by
  decide

theorem zmod9_triadic_projection :
    forall k : ZMod 9, IsResonantSubIdeal (3 * k) := by
  decide

theorem zmod9_partition_cardinality :
    (Finset.univ.filter (fun x : ZMod 9 => IsResonantSubIdeal x)).card = 3 /\
    (Finset.univ.filter (fun x : ZMod 9 => IsErrorKernel x)).card = 6 := by
  decide

-- SECTION 2: Q16.16 Fixed-Point Quadratic Recurrence & Overflow Safety
def FP_SHIFT : Nat := 16
def FP_ONE : Int := 1 <<< FP_SHIFT          -- 65,536 (1.0 in Q16.16)
def FP_HALF : Int := FP_ONE >>> 1           -- 32,768 (0.5 in Q16.16)
def ESCAPE_LIMIT : Int := 4 * FP_ONE        -- 262,144 (4.0 in Q16.16)
def INT64_MAX_VAL : Int := 9223372036854775807

def step_recurrence (zx zy ptCx ptCy : Int) : Prod Int Int :=
  let zx2 := (zx * zx) >>> FP_SHIFT
  let zy2 := (zy * zy) >>> FP_SHIFT
  let nextZx := zx2 - zy2 + ptCx
  let nextZy := ((zx * zy) >>> (FP_SHIFT - 1)) + ptCy
  (nextZx, nextZy)

def iterate_escape_fuel (ptCx ptCy : Int) (max_iter : Nat) : Nat -> (Prod Int Int) -> Nat
  | 0, _ => max_iter
  | fuel + 1, (zx, zy) =>
    let zx2 := (zx * zx) >>> FP_SHIFT
    let zy2 := (zy * zy) >>> FP_SHIFT
    if zx2 + zy2 > ESCAPE_LIMIT then
      max_iter - (fuel + 1)
    else
      let (nxtX, nxtY) := step_recurrence zx zy ptCx ptCy
      iterate_escape_fuel ptCx ptCy max_iter fuel (nxtX, nxtY)

def escape_zmod9 (ptCx ptCy : Int) : Nat :=
  iterate_escape_fuel ptCx ptCy 9 9 (0, 0)

def escape_werracle (ptCx ptCy : Int) : Nat :=
  iterate_escape_fuel ptCx ptCy 12 12 (0, 0)

lemma iterate_escape_fuel_bounded (ptCx ptCy : Int) (max_iter fuel : Nat) (z : Prod Int Int) :
    iterate_escape_fuel ptCx ptCy max_iter fuel z <= max_iter := by
  induction fuel generalizing z with
  | zero => simp [iterate_escape_fuel]
  | succ f ih =>
    cases' z with zx zy
    dsimp [iterate_escape_fuel]
    split
    . exact Nat.sub_le max_iter (f + 1)
    . exact ih _

theorem escape_zmod9_bounded (ptCx ptCy : Int) :
    escape_zmod9 ptCx ptCy <= 9 := by
  apply iterate_escape_fuel_bounded

theorem escape_werracle_bounded (ptCx ptCy : Int) :
    escape_werracle ptCx ptCy <= 12 := by
  apply iterate_escape_fuel_bounded

theorem q16_16_square_no_int64_overflow (z : Int) (h_bound : -131072 <= z /\ z <= 131072) :
    z * z <= 17179869184 /\ 17179869184 < INT64_MAX_VAL := by
  cases' h_bound with h_low h_high
  constructor
  . nlinarith
  . decide

-- SECTION 3: Non-Constant Boundary Sensitivity & Quadrant Differentiation Proof
def sigmoid_bps (x : Int) : Int :=
  let four := ESCAPE_LIMIT
  if x <= -four then 180
  else if x >= four then 9820
  else
    let mul_fp := (x * 7864) >>> FP_SHIFT
    let p := FP_HALF + mul_fp
    let res := (p * 10000) >>> FP_SHIFT
    if res < 100 then 100
    else if res > 9900 then 9900
    else res

def evaluate_fractal_bias (cx cy step : Int) : Int :=
  let e1 : Int := Int.ofNat (escape_zmod9 (cx + step) (cy + step))
  let e2 : Int := Int.ofNat (escape_zmod9 (cx - step) (cy + step))
  let e3 : Int := Int.ofNat (escape_zmod9 (cx - step) (cy - step))
  let e4 : Int := Int.ofNat (escape_zmod9 (cx + step) (cy - step))
  let w1 := ((e1 * FP_ONE) / 9) - FP_HALF
  let w2 := ((e2 * FP_ONE) / 9) - FP_HALF
  let w3 := ((e3 * FP_ONE) / 9) - FP_HALF
  let w4 := ((e4 * FP_ONE) / 9) - FP_HALF
  (w1 + w2 - w3 - w4) >>> 2

theorem escape_zmod9_non_constant_spectrum :
    escape_zmod9 140000 0 = 1 /\
    escape_zmod9 65536 65536 = 2 /\
    escape_zmod9 32768 40000 = 4 /\
    escape_zmod9 24576 48192 = 5 /\
    escape_zmod9 8192 48192 = 7 /\
    escape_zmod9 24576 31808 = 8 /\
    escape_zmod9 8192 31808 = 9 := by
  decide

theorem observer_horizon_quadrant_sensitivity :
    evaluate_fractal_bias 16384 40000 8192 != 0 /\
    (sigmoid_bps (-65536 + evaluate_fractal_bias 16384 40000 8192) >= 5000) !=
    (sigmoid_bps (65536 + evaluate_fractal_bias 16384 40000 8192) >= 5000) := by
  decide

-- SECTION 4: Parametric EVM Gas Cost Model & Sub-Ceiling Verification
def evm_gas_cost (n_grid max_iter : Nat) : Nat :=
  21000 + 100 + (n_grid * max_iter * 7) + 113

theorem evm_gas_parametric_bound (k : Nat) (h_iter : k <= 12) :
    evm_gas_cost 16 k <= 22557 /\ 22557 <= 24000 := by
  dsimp [evm_gas_cost]
  omega

end WerracleProof
\end{lstlisting}

\subsection{Summary of Machine-Verified Theorems and Axiom Audit}
Executing \texttt{lake env lean Verify.lean} (\texttt{\#print axioms}) confirms that all ten theorems rely solely on standard Lean~4 foundational axioms (with \texttt{escape\_zmod9\_non\_constant\_spectrum} requiring \emph{zero} axioms, proven purely by kernel computation):
\begin{enumerate}[leftmargin=1.6em,itemsep=2pt,topsep=2pt]
    \item \textbf{Algebraic Closure over \texttt{ZMod 9} (\texttt{zmod9\_resonant\_*}):} Proves that $\mathcal{I}_3 = \{0,3,6\}$ is an additive subgroup and multiplicative ideal of $\mathbb{Z}/9\mathbb{Z}$, that $3k \in \mathcal{I}_3$ for all $k \in \mathbb{Z}/9\mathbb{Z}$, and that $\mathbb{Z}/9\mathbb{Z}$ partitions into $|\mathcal{I}_3|=3$ resonant states and $|\mathcal{K}_{\text{error}}|=6$ non-dissipative states.
    \item \textbf{Universal Halting \& Exact Parity with \texttt{WerrMath.sol} (\texttt{escape\_*\_bounded}):} Note that in Version~1, the base case \texttt{| 0, \_ => 0} conflated non-escaping orbits (\texttt{fuel = 0}) with immediate step-0 escape (\texttt{9 - 9 = 0}). In Version~2, \texttt{| 0, \_ => max\_iter} matches line~60 of \texttt{WerrMath.sol} (\texttt{return maxIter;}), mapping escaping orbits to $\{0, \dots, N_{\max}-1\}$ and bounded interior orbits to $N_{\max}$. Structural induction proves $\text{escape\_zmod9}(c_x, c_y) \le 9$ and $\text{escape\_werracle}(c_x, c_y) \le 12$ for all $(c_x, c_y) \in \mathbb{Z}^2$.
    \item \textbf{$\mathbb{Q}_{16.16}$ Overflow Safety (\texttt{q16\_16\_square\_no\_int64\_overflow}):} Proves that whenever $|z| \le 2.0$ in $\mathbb{Q}_{16.16}$ ($|z| \le 131,072$), the unshifted square satisfies $z^2 \le 17,179,869,184 < 2^{63}-1$, guaranteeing that \texttt{int128(zx) * int128(zx)} in \texttt{WerrMath.sol} never overflows before the escape guard $|z|^2 > 4.0$ terminates the loop.
    \item \textbf{Non-Constant Spectrum \& Quadrant Sensitivity:} Theorems \texttt{escape\_zmod9\_non\_constant\_spectrum} and \texttt{observer\_horizon\_quadrant\_sensitivity} prove with zero custom axioms that the 9-step $\mathbb{Q}_{16.16}$ kernel evaluates to $1, 2, 4, 5, 7, 8, 9$ across distinct boundary points, and that at $(c_x = 16384, c_y = 40000, \Delta = 8192)$, the four quadrants yield $(5, 7, 9, 8)$, producing a non-zero fractal bias and distinct boolean decision outputs under input variation.
    \item \textbf{Parametric EVM Gas Bound (\texttt{evm\_gas\_parametric\_bound}):} Models EVM execution cost as a function of grid size $N_{\text{grid}}$ and iteration depth $k \le 12$, proving $\text{evm\_gas\_cost}(16, k) \le 22,557 \le 24,000$.
\end{enumerate}

\section{Bare-Metal 40-Core Gauntlet Evaluation}

All empirical benchmarks were executed on a dedicated bare-metal server (Dual Intel Xeon E5-2630 v4, 20 physical cores / 40 hardware threads, 256~GB DDR4 ECC RAM, Ubuntu 24.04.5 LTS, Linux kernel 6.8.0-78-generic, CPU governor locked to \texttt{performance}). The cryptographic execution seal is archived in \texttt{OED\_40CORE\_GAUNTLET\_SEAL.json} (SHA-256: \texttt{94ddaefb...061550}).

\begin{table}[htbp]
\centering
\caption{Bare-Metal 40-Core Gauntlet Telemetry (Dual Xeon E5-2630 v4, 100,000 Parallel Decisions).}
\label{tab:gauntlet}
\resizebox{\linewidth}{!}{%
\begin{tabular}{@{}lrrr@{}}
\toprule
\textbf{Benchmark Metric / Protocol} & \textbf{WERR / OED (40-Core)} & \textbf{Unvectorized Baseline (40-Core)} & \textbf{Relative Ratio / Note} \\
\midrule
Total Evaluated Decisions ($N$) & $100,000$ & $100,000$ & $40 \times 2,500$ worker chunks \\
Total Wall-Clock Execution Time & $\mathbf{6.49\text{ s}}$ & $98.40\text{ s}$ & $\mathbf{15.15\times\text{ speedup}}$ \\
Aggregate Cluster Throughput & $\mathbf{15,397.4\text{ dec/s}}$ & $1,016.3\text{ dec/s}$ & $\mathbf{15.15\times\text{ higher}}$ \\
Per-Worker Mean Latency & $\mathbf{2.337\text{ ms}}$ & $36.90\text{ ms}$ & $32\times 32$ grid, $N_{\max}=36$ \\
Per-Worker Median Latency (P50) & $\mathbf{2.349\text{ ms}}$ & $35.40\text{ ms}$ & Sub-reflex threshold \\
Per-Worker 99th Percentile (P99) & $\mathbf{2.413\text{ ms}}$ & $48.20\text{ ms}$ & Jitter $\sigma < 0.05\text{ ms}$ \\
Persistent Tensor Weight VRAM / RAM & $\mathbf{0\text{ Bytes}}$ & $528,384\text{ Bytes}$ & 3-layer dense MLP reference \\
Coordinate Seed Footprint ($\Theta$) & $\mathbf{24\text{ Bytes}}$ & --- & $3 \times \text{Float64}$ (\texttt{bytes32} on EVM) \\
\midrule
\textbf{Classification Accuracy Benchmarks} & & & \\
Two-Moons Manifold ($K=10$, Phase I \cite{dagli2026mandelbrot}) & $\mathbf{99.30\% \pm 0.18\%}$ & $99.50\% \pm 0.15\%$ & $N=1,000$, noisy non-convex \\
Two-Moons Zero-Leakage ($K=5$, OED \cite{dagli2026oed}) & $\mathbf{77.67\% \pm 5.35\%}$ & $85.67\% \pm 5.35\%$ & Pure feedforward test split \\
JevBench v1.4 Multi-Domain Triage \cite{dagli2026werr} & $\mathbf{93.10\%\text{ (Macro)}}$ & $31.80\%\text{ (Random)}$ & $326$ typed edge decisions \\
\midrule
\textbf{Phase III OED Resilience Sub-Tests} & & & \\
Cauchy L\'{e}vy-Flight Saddle Escape ($1,000$ trials) & $\mathbf{86.90\%}$ ($20.22\ \mu\text{s}$) & $12.40\%$ (Gaussian SGD) & $\|\nabla f\|<0.05, \gamma=0.12, R\ge 0.08$ \\
$M_{\text{CD4}}$ Adversarial Spike Suppression ($10^4$ pkts) & $\mathbf{100.00\%}$ & --- & $s_{\text{toxic}} \sim \mathcal{U}(2.5, 6.0)$ attenuated \\
$M_{\text{CD4}}$ Clean Signal Preservation (True Negative) & $\mathbf{89.60\%}$ ($\text{FPR}=10.4\%$) & --- & $s_{\text{clean}} \sim \mathcal{N}(0, 0.15^2)$, $8.59\text{M pkts/s}$ \\
\bottomrule
\end{tabular}%
}
\end{table}

Table~\ref{tab:gauntlet} resolves the metric reporting ambiguities of Version~1:
\begin{itemize}[leftmargin=1.6em,itemsep=2pt,topsep=2pt]
    \item \textbf{Throughput--Duration Consistency:} Across 40 worker processes ($2,500$ decisions/worker, $32 \times 32$ grid, $N_{\max}=36$ iterations per pixel), the vectorized OED kernel completes $100,000$ non-linear feature-to-logit decisions in $6.4946\text{ s}$, yielding an exact cluster throughput of $100,000 / 6.4946 = 15,397.4\text{ decisions/s}$. Compared against an unvectorized 40-core baseline ($98.40\text{ s}$, corresponding to $100,000 / 98.40 = 1,016.3\text{ decisions/s}$), both wall-clock time and throughput reflect the exact same $15.15\times$ acceleration.
    \item \textbf{Explicit Outlier Gate Specificity:} A trivial zeroing filter could achieve $100\%$ pathogen suppression by dropping all traffic. As recorded in \texttt{OED\_40CORE\_GAUNTLET\_SEAL.json}, the sigmoidal $M_{\text{CD4}}$ gate ($\gamma_{\text{steep}}=7.0, \tau_{\text{tol}}=0.45$) suppresses $100.00\%$ of high-amplitude adversarial spikes ($\mathcal{U}(2.5, 6.0)$) while preserving $\mathbf{89.60\%}$ of clean baseline Gaussian signals ($\mathcal{N}(0, 0.15^2)$) at $8,596,523\text{ packets/s}$.
\end{itemize}

\section{Domain Boundary Ablation: Discrete Rings ($\mathbb{Z}/N\mathbb{Z}$) vs.\ Continuous Manifolds}

By algebraic definition, iterating the quadratic polynomial $z_{n+1} = z_n^2 + c \pmod N$ over a finite commutative ring $\mathbb{Z}/N\mathbb{Z}$ is the foundational recurrence of Pollard's $\rho$ algorithm \cite{pollard1975} and Brent's cycle-finding improvement \cite{brent1980}. To empirically demonstrate why Phase~III OED's heavy-tailed Cauchy perturbations are specifically designed for continuous non-convex optimization rather than discrete modular cycle detection, we conducted a controlled 3-arm ablation across $180$ semi-primes $N = p \cdot q$ ($60$ trials each at $40$-bit, $48$-bit, and $56$-bit scales, evaluated in \texttt{ablation\_experiment\_3arms.py}).

\begin{table}[htbp]
\centering
\caption{Three-Arm Domain Delineation Ablation over 180 Semi-Primes ($N = p \cdot q$, 60 Trials per Bit Scale).}
\label{tab:ablation}
\resizebox{\linewidth}{!}{%
\begin{tabular}{@{}lrrr@{}}
\toprule
\textbf{Semi-Prime Scale} & \textbf{Arm 1: Classical Pollard--Brent} & \textbf{Arm 2: Phase I Seeded Base} & \textbf{Arm 3: Phase III OED (Cauchy Jumps)} \\
\midrule
40-Bit ($N = 60$ trials) & $100.0\%$ ($4,210$ mean steps) & $100.0\%$ ($4,120$ mean steps) & $0.0\%$ (Periodic cycle broken) \\
48-Bit ($N = 60$ trials) & $100.0\%$ ($12,890$ mean steps) & $100.0\%$ ($12,450$ mean steps) & $0.0\%$ (Periodic cycle broken) \\
56-Bit ($N = 60$ trials) & $100.0\%$ ($22,789$ mean steps) & $100.0\%$ ($22,341$ mean steps) & $0.0\%$ (Periodic cycle broken) \\
\midrule
\textbf{Grand Mean ($180$ trials)} & $\mathbf{100.0\%}$ ($\mathbf{13,296.3\text{ steps}}$) & $\mathbf{100.0\%}$ ($\mathbf{12,970.3\text{ steps}}$) & $\mathbf{0.0\%}$ (\textbf{Aperiodic escape}) \\
\bottomrule
\end{tabular}%
}
\end{table}

As shown in Table~\ref{tab:ablation}, Arm~2 (Phase~I unperturbed quadratic recurrence modulo $N$) behaves identically to classical Pollard--Brent (Arm~1), factoring $100\%$ of semi-primes with a grand mean of $12,970.3$ steps across all $180$ trials. Conversely, injecting Phase~III Cauchy jumps (Arm~3) every $1,000$ steps resets the modular congruence trajectory before birthday-paradox collisions can accumulate in $\gcd(\prod |x_i - y_i|, N)$. This negative result rigorously delineates the operational scope of the architecture: unperturbed deterministic recurrences govern discrete algebraic rings ($\mathbb{Z}/N\mathbb{Z}$ and $\mathbb{Z}/9\mathbb{Z}$), whereas stochastic L\'{e}vy-flight perturbations apply strictly to continuous non-convex manifolds.

\section{On-Chain EVM Execution and Verification Landscape Comparison}

In the production \texttt{Werracle.sol} oracle and the Uniswap~v4 \texttt{WerracleFeeHook.sol} contract \cite{dagli2026werracle}, the $\mathbb{Q}_{16.16}$ kernel executes natively inside EVM bytecode. Across a 1,000-vector deterministic Foundry audit battery (\texttt{SEAL\_MANIFEST.json}), standalone \texttt{decideNoul} inference averages $\mathbf{21,438\text{ gas}}$, while the Uniswap~v4 \texttt{beforeSwap()} dynamic fee hook executes at a worst-case ceiling of $\mathbf{22,557\text{ gas}}$ (strictly within the $24,000$ gas bound proven in Theorem~5). Rather than claiming complete elimination of Loss-Versus-Rebalancing (LVR) \cite{milionis2022}---which arises fundamentally from arbitrage against stale block prices---the on-chain hook \emph{mitigates} LVR and toxic order-flow exposure by dynamically modulating pool swap fees between $5\text{ bps}$ ($0.05\%$) and $50\text{ bps}$ ($0.50\%$) intra-block without external oracle latency.

\begin{table}[htbp]
\centering
\caption{Comparison of Formal Neural Verification Frameworks and On-Chain Inference Paradigms.}
\label{tab:comparison}
\resizebox{\linewidth}{!}{%
\begin{tabular}{@{}lllll@{}}
\toprule
\textbf{Architecture / Verifier} & \textbf{Verification Engine} & \textbf{Target Model Class} & \textbf{Persistent Weights} & \textbf{Native EVM Execution} \\
\midrule
Reluplex / Marabou \cite{katz2017, katz2019} & SMT Solver (NP-Complete) & Piecewise-Linear ReLU & $\mathcal{O}(W)$ Float Tensors & Infeasible ($>10^8$ gas) \\
$\alpha,\beta$-CROWN \cite{wang2021} & Linear Bound Propagation & Deep Feedforward / CNN & $\mathcal{O}(W)$ Float Tensors & Infeasible ($>10^8$ gas) \\
EZKL / ZK-ML (SNARKs) & Halo2 Arithmetic Circuit & Quantized MLP / ConvNet & Off-chain Prover RAM & $250\text{k}$--$500\text{k}$ gas (verif.) \\
\textbf{Werracle / WERR (Ours)} & \textbf{Lean 4 + Mathlib4 (0 \texttt{sorry})} & \textbf{$\mathbb{Q}_{16.16}$ Boundary Kernel} & \textbf{24-Byte Seed ($\mathcal{O}(1)$)} & \textbf{21,438--22,557 gas (native)} \\
\bottomrule
\end{tabular}%
}
\end{table}

\section{Conclusion}

By unifying two-tier procedural boundary synthesis with interactive theorem proving in Lean~4, we have demonstrated that non-linear decision kernels can be synthesized from a 24-byte coordinate seed with zero persistent tensor memory, verified free of $\mathbb{Q}_{16.16}$ arithmetic overflow, proven to exhibit non-constant quadrant sensitivity along the active escape locus, and executed natively inside EVM smart contracts below $22,557$ gas.

\section*{Statements and Declarations}

\noindent\textbf{Data and Code Availability:} All source files, Lean~4 project configurations (\texttt{WerracleProof.lean}, \texttt{Verify.lean}, \texttt{lakefile.toml}, \texttt{lean-toolchain}), Solidity smart contracts (\texttt{WerrMath.sol}, \texttt{Werracle.sol}, \texttt{WerracleFeeHook.sol}), 40-core benchmark scripts (\texttt{benchmark\_oed\_40cores.py}, \texttt{ablation\_experiment\_3arms.py}), and cryptographic manifests (\texttt{OED\_40CORE\_GAUNTLET\_SEAL.json}) are openly archived on CERN Zenodo (\href{https://doi.org/10.5281/zenodo.22974544}{\texttt{doi:10.5281/zenodo.22974544}}) and GitHub (\url{https://github.com/pCwOrM/werracle}, \url{https://github.com/pCwOrM/werr}, \url{https://github.com/pCwOrM/mandelbrot-fractal-neural-synthesis}).

\vspace{0.4ex}
\noindent\textbf{Competing Interests:} Volkan Dağlı is affiliated with ITouch Systems and is the applicant/co-inventor, together with Zerrin Dağlı and Dağhan Dağlı, on Turkish Patent and Trademark Office (T\"{U}RKPATENT) national priority patent application No.~TR~2026/016285 covering procedural weight derivation and on-chain decision synthesis architectures.

\vspace{0.4ex}
\noindent\textbf{Declaration of Generative AI in Scientific Writing:} In accordance with COPE and international publishing ethics guidelines, the authors declare that generative AI tools were utilized strictly to assist with LaTeX mathematical typesetting, Lean~4 proof script structuring, and English language editing. All theoretical formulations, experimental designs, bare-metal server executions, and final manuscript verifications were conducted and validated by the authors.

\end{document}